\documentclass[preprint,10pt]{elsarticle}

\usepackage{float}
\usepackage[numbers]{natbib}
\usepackage{mathtools}
\usepackage{url}
\usepackage{hyperref}
\usepackage{mathtools}
\usepackage{amssymb}
\usepackage{hyperref}
\usepackage{xcolor}
\usepackage{mdframed}
\usepackage{xcolor}
\usepackage{mdframed}
\usepackage{array}
    \newcolumntype{C}[1]{>{\centering\arraybackslash}p{#1}}

\begin{document}
\begin{frontmatter}
\title{Cross-Breed Pig Identification Using Auricular Vein Pattern Recognition: A Machine Learning Approach for Small-Scale Farming Applications}

\author[cmu]{Emmanuel Nsengiyumva}
\author[cmu]{Leonard Niyitegeka}
\author[cmu]{Eric Umuhoza}

\address[cmu]{Carnegie Mellon University Africa, Kigali, Rwanda\\ Email:{eumuhoza@andrew.cmu.edu}}

\begin{abstract}
Accurate livestock identification is a cornerstone of modern farming: it supports health monitoring, breeding programs, and productivity tracking. However, common pig identification methods, such as ear tags and microchips, are often unreliable, costly, target pure breeds, and thus impractical for small-scale farmers. To address this gap, we propose a noninvasive biometric identification approach that leverages  uniqueness of the auricular vein patterns. To this end, we have collected 800 ear images from 20 mixed-breed pigs (Landrace \verb|cross| Pietrain and Duroc \verb|cross| Pietrain), captured using a standard smartphone and simple back lighting. A multistage computer vision pipeline was developed to enhance vein visibility, extract structural and spatial features, and generate biometric signatures. These features were then classified using machine learning models. Support Vector Machines (SVM) achieved the highest accuracy: correctly identifying pigs with 98.12\% precision across mixed-breed populations. The entire process from image processing to classification was completed in an average of 8.3 seconds, demonstrating feasibility for real-time farm deployment.
We believe that by replacing fragile physical identifiers with permanent biological markers, this system provides farmers with a cost-effective and stress-free method of animal identification. More broadly, the findings confirm the practicality of auricular vein biometrics for digitizing livestock management, reinforcing its potential to extend the benefits of precision farming to resource-constrained agricultural communities.
\end{abstract}

\begin{keyword}
Pig identification \sep Auricular veins \sep Biometrics \sep Smart farming \sep Computer vision \sep Machine learning
\end{keyword}

\end{frontmatter}

\section{Introduction}
Accurate individual identification of livestock represents a fundamental requirement for modern agricultural practices, particularly in pig farming: individual identification of pigs allows farmers to implement targeted health monitoring protocols, maintain comprehensive breeding records, optimize feeding strategies based on individual growth patterns, and ensure complete traceability throughout the production cycle \cite{kapun2020uhf,fuentes2022ai}. These capabilities become increasingly critical as global food security demands intensify and agricultural systems transition toward precision farming approaches that maximize resource efficiency while maintaining animal welfare standards.

The emergence of smart farming technologies has catalyzed significant interest in biometric identification systems for livestock management. Unlike traditional industrial farming methods that treat animals as homogeneous groups, precision livestock farming recognizes the importance of individual animal monitoring to optimize production results \cite{meng2025biometrics}.
As discussed in \cite{neethirajan2023digital2030} and \cite{himu2024digital}, there is an increasing emphasis on technologies that support real-time monitoring and data-driven decision-making in livestock farming. For small-scale farmers with limited resources, these systems must also overcome the cost and technical barriers that have historically hindered the adoption of advanced digital tools. 

The integration of computer vision and machine learning technologies into agricultural applications has opened new possibilities for non-invasive livestock identification. Biometric approaches leverage unique biological characteristics inherent to each animal, offering permanent identification solutions that eliminate the recurring costs and maintenance challenges associated with external identification devices \cite{dan2023ear,mustafi2020biometric}. These technologies align with broader agricultural digitization trends that aim to improve farm management efficiency, improve animal welfare through reduced handling stress, and enable data-driven decision-making processes essential for sustainable livestock production.

\subsection{Current approaches for pig identification}
\label{subsect:approach}
Conventional pig identification methods have predominantly relied on physical markers, including radio frequency identification (RFID) tags, ear notching, tattooing, and subcutaneous microchips\cite{ziyabari2014rfid, peng2025ear, lomax2018ear}. RFID technology, while enabling automated data collection and feeding management, presents substantial challenges for small-scale operations. The tags attached to the ears of the pigs are susceptible to damage from aggressive animal behavior, environmental exposure, and mechanical wear, resulting in the loss of identification data that disrupts critical monitoring and breeding programs \cite{kapun2020uhf,peng2025ear}. The economic burden of replacement of the tag, the requirements of specialized reading equipment, and the technical expertise necessary for system maintenance create significant barriers for farmers limited in resources \cite{rob2024rfid}.

Microchip identification, although offering greater durability than external tags, requires invasive implantation procedures and specialized scanning equipment that may be prohibitively expensive for small-scale operations. Furthermore, both RFID and microchip systems require physical proximity for data retrieval, limiting their applicability to continuous monitoring applications \cite{peng2025ear}. Ear notching and tattooing methods, while cost-effective, provide limited information storage capacity and suffer from readability degradation over time, particularly in outdoor farming environments.

Body pattern recognition methods have been explored for livestock species with distinctive coat patterns, but they are inadequate for pigs where individual morphological variations are insufficient for reliable identification \cite{meng2025biometrics}. Iris and retinal analysis, while offering exceptional accuracy for human biometrics, face practical limitations in pig applications due to anatomical constraints, specialized imaging requirements, and the need for animal restraint during capture procedures\cite{meng2025biometrics,cihan2023biometric,10748293}.

Recent advances in computer vision have led to the investigation of camera-based biometric identification approaches for livestock. Muzzle print recognition has shown promising results for cattle identification, achieving accuracies that exceed 99\% by analyzing unique ridge patterns similar to human fingerprints \cite{b15}. However, application to pigs presents significant challenges due to saliva secretion that obscures muzzle patterns and the requirement of close proximity image capture in active farm environments \cite{chakraborty2020pig}. 

Facial recognition systems can identify individual pigs based on distinctive facial features and markings \cite{Rong2023facial}. Rong Wang et al. \cite{b18} proposed a lightweight pig face recognition model using automatic detection and knowledge distillation. This method emphasizes computational efficiency and could be advantageous in resource-limited environments. Furthermore, Jeong Se Yeon et al. \cite{b19} combined YOLO for real-time detection of pig faces with a Vision Transformer (ViT) model for classification. While promising, these methods may require further validation in diverse farm environments to assess their robustness and scalability.

Beyond facial features, auricular vein-based biometrics have been explored as an alternative approach. Dan et al. \cite{9782062} investigated Ghungroo pigs (black), capturing patterns of ear veins with controlled illumination and applying machine learning techniques for classification. In a separate study, Yorkshire pigs were analyzed using auricular vein branching point templates, where recognition was performed using Euclidean distance-based matching \cite{dan2021ear}. More recently, the Auricular Ear Image Network (AEIN) framework refined this research direction by introducing structured representations of vein features and alternative matching strategies \cite{b16}.  These works demonstrate the potential of ear vein structures as stable biometric identifiers, though they remain in early research stages with evaluations limited to single breeds under controlled conditions.

A critical limitation across existing approaches is the focus on single-breed populations under controlled experimental conditions. Real-world farming operations typically maintain multiple breeds of pigs with varying morphological characteristics, age ranges, and environmental exposure patterns. The transferability and robustness of current identification systems across diverse breed compositions remain largely invalidated, representing a significant gap between research achievements and practical deployment requirements.

Furthermore, most existing biometric systems require substantial computational resources, specialized hardware, or controlled imaging conditions that may not be feasible in resource-constrained farming environments. The lack of comprehensive validation across diverse environmental conditions, breed compositions, and age ranges limits the practical applicability of current approaches for widespread adoption in small-scale farming operations.

\subsection{Research gap}
\label{subsect:gaps}
Although various biometric approaches have shown potential for livestock identification, auricular vein pattern recognition remains largely unexplored for pig applications. The unique vascular architecture within the ears of pigs offers several theoretical advantages for biometric identification: vein patterns are uniquely unique to each individual, remain stable throughout the lifetime of the animal, and are accessible by non-invasive imaging techniques \cite{9782062,b16}. Unlike external features susceptible to environmental damage or aging effects, internal vascular structures provide consistent biometric signatures that are suitable for long-term identification applications.

Existing studies on auricular vein biometrics for pigs have been limited to single-breed populations under controlled experimental conditions. Early work by Dan et al. \cite{9782062} demonstrated that ear vein structures of Ghungroo pigs could be effectively captured and classified using feature extraction and machine learning techniques. This was followed by the AEIN approach \cite{b16}, which refined the methodology by emphasizing the branches of the vein and testing alternative matching strategies. Although both studies highlighted the promise of auricular vein features, they remain constrained by simplified feature sets and their focus on single breeds.

The absence of validated, non-invasive identification solutions specifically designed for small-scale farming environments represents a critical gap in current research. Existing commercial solutions typically require expensive specialized equipment, controlled imaging conditions, or extensive technical expertise that may not be available to resource-constrained farmers. This technology gap restricts access to precision farming benefits that could greatly increase productivity and sustainability.

\subsection{Our contributions}
The contributions of this study are twofold.

\begin{enumerate}
    \item A feature extraction technique for capturing the distinctive characteristics of the ear veins. The proposed methodology involves the complete extraction of features from venous networks, the acquisition of structural details, spatial relationships, branching patterns and geometric properties to build detailed biometric representations.
    \item A lightweight cross-breed identification system based on ear vein patterns. The proposed solution demonstrates high performance ( 98.12\% precision) in mixed populations, addressing the critical gap in cross-breed pig identification observed in state-of-the-art approaches. In addition, such high precision is achieved while maintaining computational efficiency suitable for smartphone deployment: a viable solution for small-scale farmers.  
\end{enumerate}

The remainder of this paper is organized as follows: In Section \ref{sec:methodology}, we present our approach, covering data collection, ear vein feature extraction, to machine learning classification for individual pig identification. Section  presents the comprehensive methodology, including dataset collection procedures, the multi-stage computer vision pipeline for vein feature extraction, and machine learning classification approaches. Section \ref{sec:results} evaluates the effectiveness of the proposed feature extraction techniques, assesses the reliability of ear vein patterns across different breeds of pigs, and compares the robustness of various machine learning algorithms for practical implementation. Section \ref{sec:conclusions} concludes the study by summarizing key contributions and highlighting the potential for auricular vein biometrics in precision livestock farming applications.
\label{subsect:contributions}

\section{Methodology}
\label{sec:methodology}
This research presents a biometric identification system for pigs based on the recognition of ear vein patterns. The proposed approach employs computer vision techniques combined with machine learning algorithms to create unique biometric templates from pig ear vascular patterns. The system addresses the limitations of traditional identification methods by providing a noninvasive, cost-effective, and highly accurate solution suitable for small-scale farming environments.

As depicted in Figure 1, the proposed system follows a multistage pipeline designed to process images of the pig ear and generate discriminative biometric features. The architecture consists of extraction of the region of interest (as presented in Subsection \ref{subsec:roi}), extraction of vein characteristics (as presented in Subsection \ref{subsec:feature extraction}), generation of feature vectors, and ML classification algorithms (as presented in Subsection \ref{subsec:edge detect}).

\subsection{Dataset and experimental setup}

The dataset used in this study comprises 800 ear images collected from 20 pigs, including 15 Landrace \verb|cross| Pietrain and 5 Duroc \verb|cross|  Pietrain, aged between 4 and 6 months, with each pig contributing exactly 40 images.

The image acquisition process was designed to highlight vascular patterns within the ear using a 12 MP phone camera (4032 × 3024 pixels) under controlled lighting conditions. Each pig was placed in a relatively dark environment to minimize ambient light interference, while a direct light source from a simple torch, positioned beneath the ear with the help of the farm security team, illuminated the ear tissue from below. The images were captured from the opposite side, creating a translucent effect in which red pigmentation dominated and the vein structures appeared as dark thread-like patterns. The 12 MP resolution provided adequate detail for extracting fine vein structures while ensuring computational efficiency, with the camera's sensor quality directly affecting the subsequent red-channel-based vein detection performance. This data collection method ensured consistent image quality between subjects and improved the visibility of vein patterns, which are primary biometric trait for subsequent feature extraction and classification.

To allow a balanced and unbiased evaluation, the dataset was partitioned using stratified sampling, 80\% (i.e 640 images while using both breeds) were allocated for training and 20\% for testing (i.e 160 images while using both breeds), with equal representation of each pig in both subsets. This careful preparation ensured a reliable assessment of the performance and generalization of the proposed pig ear vein identification system


\begin{figure}[H]
    \centering
    \includegraphics[width=\linewidth]{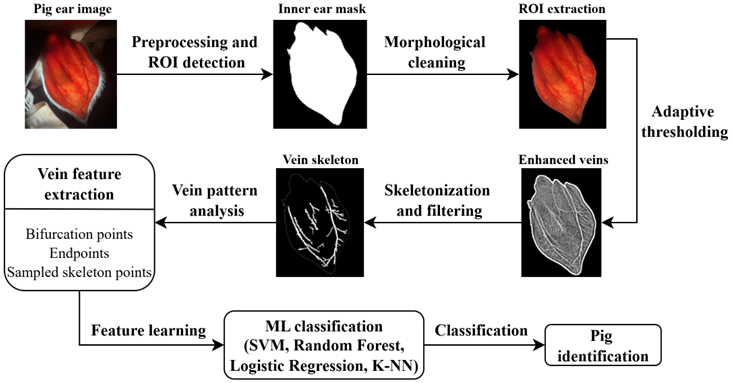}
    \caption{Complete pipeline architecture for pig identification using auricular vein patterns, illustrating the sequential processing stages from raw ear image acquisition through ROI extraction, vein feature extraction, feature vector generation, to final machine learning classification for individual pig identification.}
    \label{fig:architecture}
\end{figure}


\subsection{Region of interest (ROI) extraction}
\label{subsec:roi}

The ROI extraction stage isolates the inner region of the pig's ear, providing a clean area for subsequent vein feature analysis. This is achieved by leveraging the dominance of the red color channel, as veins and inner ear structures appear more prominently in red under illumination. 

By examining the images, it was evident that the pigs' ears, our area of interest, were most prominent in the red channel. To isolate this area, we set thresholds for the red channel, the red-to-green ratio, and the red-to-blue ratio. Given the variation in intensity, brightness, and contrast of pixels in the images, we explored several adaptive thresholding methods, as using a fixed threshold would not have been effective.

Of the methods we tried--- the Otsu thresholding method \cite{otsu_thresholding}, statistical thresholding \cite{MAKSYMENKO201795}, and adaptive Gaussian thresholding\cite{gaussian_adaptive_thresholding}--- to automatically determine the optimal cutoff values for the red channel intensity, as well as the red-to-green and red-to-blue ratios in each image, none were found to be suitable for our dataset. Therefore,  we developed a custom adaptive thresholding pipeline to segment the inner ear region from RGB images of pigs, designed to adapt to variations in illumination, contrast, and skin pigmentation.

Let $R$, $G$, and $B$ denote the red, green, and blue channels of the image. Two ratio maps are computed for each pixel:
\[
\text{R/G} = \frac{R}{G + \epsilon}, \quad \text{R/B} = \frac{R}{B + \epsilon},
\]
where $\epsilon$ is a small constant to avoid division by zero. The red channel threshold $T_R$ is computed adaptively based on the standard deviation $\sigma_R$ of the red channel:
\begin{itemize}
    \item for low-contrast images ($\sigma_R < 10$), $T_R$ is set using a percentile of the red channel values, clamped between 20 and 35;
    \item for medium-contrast images ($10 \leq \sigma_R < 30$), the threshold is similarly percentile-based;
    \item for high-contrast images ($30 \leq \sigma_R < 60$), $T_R$ is computed from the mean-normalized red values $(R - \mu_R)/\sigma_R$ scaled by 0.3 and clamped between 20 and 35; and
    \item for very high-contrast images ($\sigma_R \geq 60$), the scaling factor is 0.5, also clamped to the same range.
\end{itemize}
The bounds for all those intervals were determined by experimentation with our dataset, taking into account the appropriate thresholds appropriate to the conditions under which the images were captured.\\ 
The ratio thresholds $T_{\text{R/G}}$ and $T_{\text{R/B}}$ are initially set at the 90th percentile of their respective ratio distributions. Those thresholds are then adjusted based on the mean read intensity $\mu_R$: for bright images ($\mu_R > 60$), the thresholds are scaled down, while for dark images ($\mu_R < 10$), the thresholds are scaled up. Images of medium brightness undergo intermediate scaling. Finally, the inner ear mask is obtained by selecting pixels that simultaneously satisfy the following conditions:
\[
R \geq T_R, \quad \text{R/G} \geq T_{\text{R/G}}, \quad \text{R/B} \geq T_{\text{R/B}}.
\]


To remove noise and fill small gaps, morphological operations are applied. A closing operation fills small holes within the mask, while an opening operation removes isolated noise pixels. Mathematically, using a structuring element $B$ (a $5 \times 5$ square):
\[
M_2 = (M_1 \oplus B) \ominus B \quad \text{(closing)}, \quad
M_3 = (M_2 \ominus B) \oplus B \quad \text{(opening)}
\]
where $\oplus$ and $\ominus$ denote dilation and erosion \cite{opencv_morph}, respectively.

Next, connected component analysis labels all contiguous regions in $M_3$. Let $L(x, y)$ be the label of each pixel and let $S_k = \sum_{x,y} \mathbf{1}_{L(x,y)=k}$ be the size of the component $k$. The largest component 
\[
k_\text{largest} = \arg\max S_k
\] 
is selected as the inner ear, forming a binary mask $M_4$ where 
\[
M_4(x, y) = 
\begin{cases}
1, & L(x, y) = k_\text{largest}, \\
0, & \quad \text{otherwise}.
\end{cases}
\]

To ensure a solid mask, any holes in $M_4$ are filled by tracing and filling its external contours, resulting in the final mask $M_\text{filled}$. Finally, this mask is applied to the original RGB image:
\[
I_\text{ROI}(x, y) = I_\text{RGB}(x, y) \cdot M_\text{filled}(x, y)
\]

This produces a masked image that contains only the inner-ear region, along with its corresponding binary mask.

\begin{mdframed}[
  leftmargin=0,
  rightmargin=0,
  skipabove=\topsep,
  skipbelow=\topsep
  ]
 By combining Otsu-based thresholding, morphological refinement, connected component analysis, and hole filling, the procedure ensures a clean and precise ROI suitable for vein feature extraction and reliable biometric recognition.
\end{mdframed}

\begin{figure}[H]
    \centering
    \includegraphics[width=\linewidth]{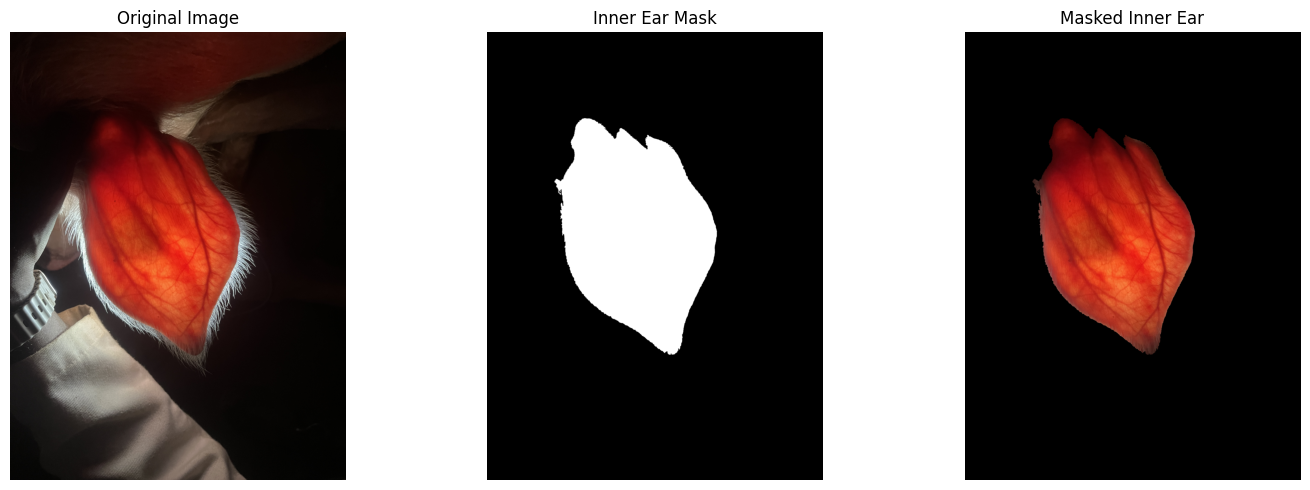} 
    \caption{ROI extraction process showing: (a) original ear image, (b) refined mask after morphological operations, and (c) final masked inner ear region for vein analysis.}
    \label{roi}
\end{figure}

\subsection{Vein feature extraction}
\label{subsec:feature extraction}
The extraction of vein characteristics is a critical step in the pig identification system, as vein patterns in the inner ear provide unique biometric information. The extraction process combines image processing, morphological operations, and skeleton-based feature analysis to obtain robust vein descriptors.

\subsubsection{Input and preprocessing}

Let $I \in \mathbb{R}^{H \times W \times 3}$ denote the RGB image of the masked inner ear and $M \in \{0,1\}^{H \times W}$ be the binary mask of the inner ear region. The red, green, and blue channels were first extracted:

\[
I = [R, G, B], \quad R, G, B \in \mathbb{R}^{H \times W}.
\]

Since veins appear darker in the red channel, the red channel is inverted:

\[
R_{\text{inv}} = 255 - R.
\]

The contrast is enhanced using \textit{ contrast-limited adaptive histogram equalization (CLAHE)\cite{clahe}}:

\[
R_{\text{enhanced}} = \text{CLAHE}(R_{\text{inv}}),
\]

which improves the local contrast and highlights vein structures.

\subsubsection{Image sharpening}

The edges of the veins are emphasized by applying a sharpening kernel via convolution:

\[
K = 
\begin{bmatrix}
0 & -1 & 0 \\
-1 & 5 & -1 \\
0 & -1 & 0
\end{bmatrix}, 
\quad 
R_{\text{sharpened}} = R_{\text{enhanced}} * K,
\]

where $*$ denotes 2D convolution and the K kernel was chosen for its strong ability to sharpen the images, which in turn helps the model detect more veins accurately. 
\subsubsection{Adaptive thresholding}

The image is converted to a binary representation using adaptive thresholding.

\[
R_{\text{binary}}(x,y) =
\begin{cases}
1 & \text{if } R_{\text{sharpened}}(x,y) > T(x,y) \\
0 & \text{otherwise}
\end{cases},
\]

where $T(x,y)$ is the local mean of the neighborhood around the pixel $(x,y)$. This step separates the veins from the background under varying illumination.

\subsubsection{Morphological cleaning}

Small noise is removed using a morphological opening with a $3 \times 3$ structuring element $S$:

\[
R_{\text{clean}} = R_{\text{binary}} \circ S,
\]

 small objects below a minimum size threshold (e.g., 300 pixels) are discarded:

\[
R_{\text{filtered}} = \text{remove\_small\_objects}(R_{\text{clean}}, \text{min\_size}=300).
\]

To connect fragmented veins, dilation is applied:

\[
R_{\text{dilated}} = R_{\text{filtered}} \oplus S.
\]

\subsubsection{Connected component analysis}

Connected components are labeled and only the largest $N$ components are retained, that is, \[
\{C_1, C_2, \dots, C_N\} \subset R_{\text{dilated}}.
\]

This ensures that only significant vein structures are considered.

\subsection{Edge detection and final cleaning}
\label{subsec:edge detect}

The edges of the major veins in the image are identified using the \textit{Canny edge detector}. This is a widely used method for detecting edges in images because it efficiently finds areas where the intensity of the pixels changes rapidly. A key advantage of the Canny detector is that, while it is very effective in finding edges, it also uses less memory compared to other edge detection methods such as the Sobel algorithm \cite{canny}.

After detecting the edges, the result is combined with the original major vein regions to connect segments that are close to each other. This ensures that the vein structures are more complete and continuous.

Finally, a mask representing the inner ear region is applied to the processed image. This step ensures that only the veins located within the ear are kept, removing any structures outside the area of interest.

\subsubsection{Skeletonization and feature extraction}

The vein image is skeletonized to obtain a one-pixel-wide representation \cite{Davies2003VisualInspection}.  
The features of the vein are extracted by analyzing the 3x3 neighborhood around each skeleton pixel $(x,y)$. The extracted features include endpoints, bifurcations, and sampled skeleton points, whereby endpoints are pixel with exactly one neighbor, whereas bifurcations are pixels with more than two neighbors. In addition to that, a subset of skeleton points is sampled uniformly to represent the vein's shape.
These features are robust descriptors of inner ear vein patterns, providing a unique signature suitable for classification.

\subsubsection{Feature vector generation}

After extracting the characteristics of the vein from the inner ear, the next step is to convert these characteristics into a standardized numerical representation suitable for classification. The generated feature vector captures the statistical, spatial and structural characteristics of the vein network. The extracted vein features include three sets of points:
\begin{enumerate}
    \item Bifurcation points $\{B_i = (x_i, y_i)\}$, where vein branches occur. 
Pairwise Euclidean distances between bifurcation points are computed as:
\[
d_{ij} = \sqrt{(x_i - x_j)^2 + (y_i - y_j)^2}, \quad i \neq j
\]
From these distances, the \textit{mean} and \textit{standard deviation} are calculated to characterize the typical spacing and variability between the branches of the vein. Additionally, the relative angles between each pair of bifurcation points are computed:
\[
\theta_{ij} = \arctan2(y_j - y_i, x_j - x_i)
\]
These angles are grouped into a histogram (e.g., 8 bins over $[- \pi, \pi]$) and normalized, providing a rotationally invariant descriptor of the vein branching orientations.\\

\item Endpoints $\{E_j = (x_j, y_j)\}$, representing terminal points of veins. 
Similarly, pairwise distances between endpoints are calculated, and their mean and standard deviation are calculated. These measures capture the overall spatial spread of the vein terminations.\\

\item Sampled skeleton points $\{S_k = (x_k, y_k)\}$, providing a uniform representation of vein pathways. 
  To capture the global geometry of the vein network, $x$ and $y$ coordinates of the skeleton points sampled are analyzed. The mean and standard deviation of each coordinate provide information about the center and dispersion of the vein pattern.
\[
x_\text{mean} = \frac{1}{N}\sum_k x_k, \quad x_\text{std} = \sqrt{\frac{1}{N}\sum_k (x_k - x_\text{mean})^2}
\]
\[
y_\text{mean} = \frac{1}{N}\sum_k y_k, \quad y_\text{std} = \sqrt{\frac{1}{N}\sum_k (y_k - y_\text{mean})^2}
\]
Additionally, a two-dimensional histogram is computed on the sampled points to create a spatial density map, which is normalized to remove scale dependency. For computational efficiency, only a subset of histogram bins can be retained in the final feature vector.
\end{enumerate}

\begin{mdframed}[
  leftmargin=0,
  rightmargin=0,
  skipabove=\topsep,
  skipbelow=\topsep
  ]
 \emph{Resulting Feature vector.}
The extracted features are concatenated into a single fixed-length representation of \emph{68 dimensions}, capturing both structural and statistical information about the inner ear vein network. Specifically, the feature vector is composed of six groups: 
\begin{enumerate}
    \item count features---2 dimensions, 
    \item bifurcation statistics---2 dimensions, 
    \item endpoint statistics---2 dimensions, 
    \item spatial distribution statistics---4 dimensions, 
    \item angle histogram---8 dimensions, and 
    \item spatial density histogram---50 dimensions.
\end{enumerate}
The count features encode the number of bifurcation and endpoint points, while the bifurcation and endpoint statistics capture the mean and variance of interpoint distances. Spatial distribution statistics summarize the positional spread of features across the image using coordinate means and standard deviations.
\end{mdframed}

The higher-dimensional components are provided by the histograms. The angle histogram, with 8 bins, encodes the directional relationships between bifurcation points, while the spatial density histogram describes the geometric distribution of the vein patterns. Although a full $10 \times 10$ histogram would yield 100 bins, only the first 50 are retained to achieve compactness while maintaining discriminative power. Together, these components form the final feature vector:

\begin{scriptsize}
  \[
\mathbf{f} =
\Big[
N_B, \; N_E, \; \mu_B, \; \sigma_B, \; \mu_E, \; \sigma_E, \;
x_{\text{mean}}, \; x_{\text{std}}, \; y_{\text{mean}}, \; y_{\text{std}}, \;
\text{angle histogram}, \; \text{spatial histogram}
\Big]
\]
\end{scriptsize}

This \emph{68-dimensional characteristic vector} provides a normalized and discriminative description of the vein network, integrating structural, spatial, and orientation information that can be effectively leveraged in the classification stage.

\subsubsection{Classification models}
To develop a lightweight and accurate model for mobile deployment, we evaluated four machine learning algorithms---\textit{Support Vector Machines (SVM)}, \textit{Random Forests (RF)}, \textit{K-Nearest Neighbors (KNN)}, and \textit{Logistic Regression (LR)}---on high-dimensional feature vectors extracted from pig ear vein patterns.

Each model was trained on standardized feature vectors and optimized to balance accuracy, efficiency, and robustness: SVM with an RBF kernel to capture nonlinear patterns, RF with an ensemble of decision trees for stability, KNN with distance-weighted voting for simplicity and effectiveness, and LR with regularization and a one-vs-rest strategy for multiclass classification.

Their performance was assessed primarily using classification accuracy, providing a reliable framework for practical identification of pigs in real world applications.

\section{Results and Discussion}
\label{sec:results}
In this section, we present our results and compare them with state-of-the-art pig identification methods. 

\subsection{Effectiveness of feature extraction techniques for capturing distinctive vein characteristics}

The effectiveness of vein-based biometric identification relies on the quality and discriminative power of the extracted characteristics. In the following paragraphs,we evaluate the proposed feature extraction methodology by examining its ability to capture distinctive vein characteristics and comparing its performance with existing approaches.

As detailed in Section \ref{sec:methodology}, the extraction process begins with the masked inner ear region and applies contrast enhancement using CLAHE, followed by image sharpening and adaptive thresholding to isolate vein structures.

\begin{figure}[H]
    \centering
    \includegraphics[width=\linewidth]{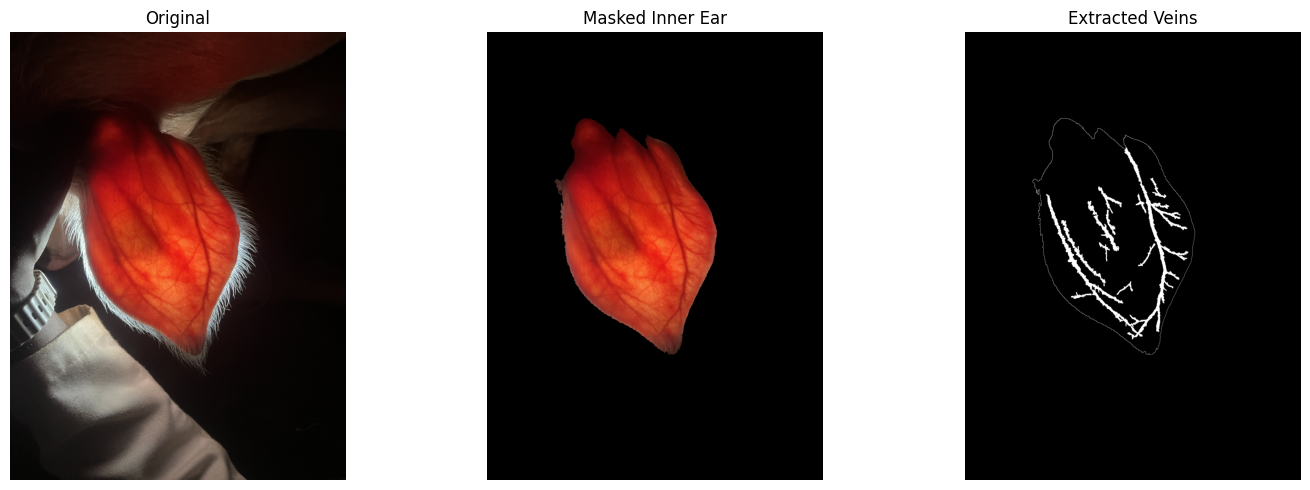}
    \caption{Vein extraction pipeline showing: (left) original ear image, (center) masked inner ear region, and (right) extracted vein patterns after processing.}
    \label{fig:vein_extraction}
\end{figure}

Figure~\ref{fig:vein_extraction} illustrates the vein extraction pipeline, showing the progression from the original image of the ear through the masked inner ear region to the final vein patterns extracted.
\begin{mdframed}[
  leftmargin=0,
  rightmargin=0,
  skipabove=\topsep,
  skipbelow=\topsep
  ]
 The process successfully isolates the complex vascular network within the ear, giving clear vein structures suitable for feature analysis.
\end{mdframed}

\subsection{Reliability of ear vein patterns for cross-breed pig identification}

To assess the consistency of ear vein patterns for cross-breed pig identification, we conducted classification experiments using three different data set configurations.
Four state-of-the-art machine learning algorithms were  evaluated: SVM with RBF kernel, Random Forest with varying estimators, K-Nearest Neighbors (k = 5) and Logistic Regression with a one-vs-rest multiclass strategy.
Each algorithm was trained using identical feature sets and evaluated under consistent conditions to ensure a fair comparison. The results, summarised in Table~\ref{tab:performance_comparison}, demonstrate that the SVM consistently outperforms the other classifiers in different configurations of the data set. 

\begin{table}[t]
\centering
\caption{Classification performance of four machine learning algorithms across three dataset configurations showing SVM's consistent superiority.}
\label{tab:performance_comparison}
\resizebox{\textwidth}{!}{
\begin{tabular}{|p{4cm}| C{1.2cm}| C{1cm}| C{2cm} |C{1.2cm}| C{3cm}|}
\hline
\textbf{Dataset Configuration} & \textbf{Sample Size} & \textbf{SVM (\%)} & \textbf{Random Forest (\%)} & \textbf{KNN (\%)} & \textbf{Logistic Regression (\%)} \\
\hline
Mixed Breeds (All Pigs) & 20 & \textbf{98.12} & 95.00 & 96.25 & 96.25 \\
Landrace-Pietrain Only & 15 & \textbf{99.17} & 95.00 & 93.33 & 93.33 \\
Duroc-Pietrain Only & 5 & 98.33 & 97.50 & 95.00 & \textbf{100.00} \\
\hline
\textbf{Average Performance} & - & \textbf{98.54} & 95.83 & 94.86 & 96.53 \\
\hline
\end{tabular}
}
\end{table}

\subsubsection{Machine learning algorithms comparison}

\paragraph{Logistic Regression} Although logistic regression achieved a perfect classification (100\%) in the smallest Duroc-Pietrain subset, this performance dramatically decreased to 93.33\% in the larger Landrace-Pietrain dataset and 96.25\% in the mixed breed configuration. The high variance, 6.67\%, indicates severe overfitting tendencies, especially with limited training samples. 
Logistic regression, as a linear classifier, struggles to capture the nonlinear structure of vein patterns, and inadequate regularization further increases the risk of overfitting\cite{friedrich2023regularization}.

\paragraph{Random Forest} Random Forest delivered robust and consistent performance (95.0-97.5\%) across all configurations with minimal 2.5\% variance, demonstrating superior stability and reliable generalization across diverse breed compositions.

\paragraph{KNN} KNN maintained  consistency comparable to RF, with only 2.92\% variance, establishing both algorithms as reliable choices for cross-breed pig identification.
Although both algorithms significantly outperformed Logistic Regression in terms of stability, SVM's ability to handle high-dimensional biometric data with complex decision boundaries made it the \textbf{optimal choice} for this application.

\begin{figure}[H]
    \centering
    \includegraphics[width=\linewidth]{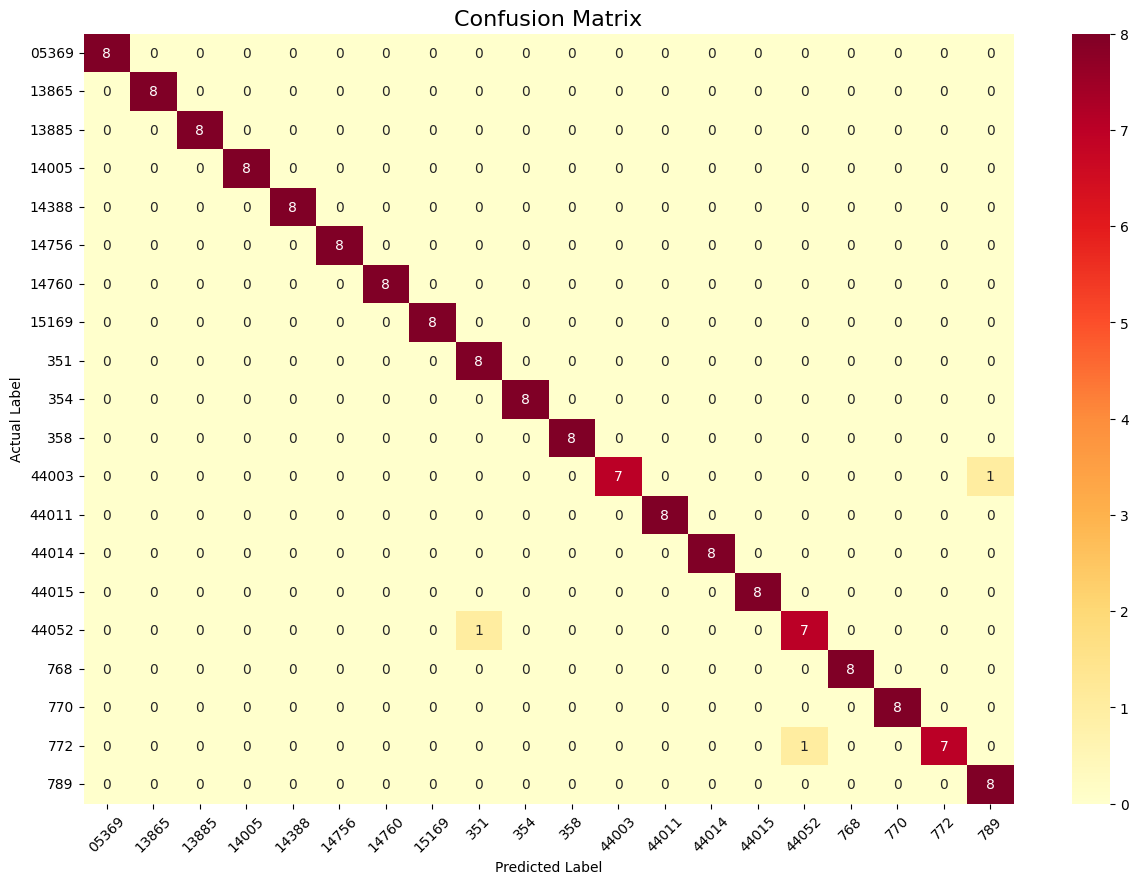}
    \caption{SVM confusion matrix for mixed-breed classification showing high diagonal accuracy with minimal off-diagonal misclassifications.}
    \label{fig:confusion_matrix}
\end{figure}

\begin{mdframed}[
  leftmargin=0,
  rightmargin=0,
  skipabove=\topsep,
  skipbelow=\topsep
  ]
 \emph{SVM} demonstrated the most consistent performance across all experimental configurations, with minimal variance, 1.05\%,  between the best and worst performance. This consistency indicates a superior generalization ability when handling complex, high-dimensional biometric features with inherent biological variations across different breeds. 
  The RBF kernel effectively captures non-linear relationships within the 68-dimensional feature space, enabling robust classification boundaries that generalize well to unseen data.

 \paragraph{Error analysis and misclassifications patterns}
Table~\ref{tab:error_analysis} shows that out of 160 test images, only 3 misclassifications occurred, both for false negatives and false positives. 
The errors were randomly distributed across breeds, with no systematic patterns observed, suggesting that the misclassifications were due to natural individual variation rather than breed-specific or algorithmic bias. The confusion matrix in Figure~\ref{fig:confusion_matrix} shows a strong diagonal trend, indicating that most pigs were correctly identified.
The few errors observed reflect instances where similar vein patterns led to confusion between individuals.

\end{mdframed}



\begin{table}[t]
\centering
\caption{Error distribution analysis for SVM classification showing conservative behaviour with only three misclassified images on mixed breeds.}
\label{tab:error_analysis}
\resizebox{\textwidth}{!}{
\begin{tabular}{|p{3.3cm}| C{1cm}| C{2.8cm}| C{3.3cm}| C{3.5cm}|}
\hline
\textbf{Error type} & \textbf{Count} & \textbf{Percentage (\%)} & \textbf{Breed distribution} & \textbf{Primary Cause} \\
\hline
False negatives & 3 & 1.88 & 2 Landrace--pietrain 1 Duroc--pietrain & Vein pattern similarity \\
\hline
False positives & 3 & 1.88 & 2 Landrace--pietrain 1 Duroc--pietrain & Vein pattern similarity \\
\hline
Total Errors & 6 & 3.76 & Mixed & Individual variation \\
\hline
Correctly classified & 157 & 98.12 & Mixed & Robust features\\
\hline
\end{tabular}
}
\end{table}


\subsubsection{System performance and deployment feasibility}

The proposed system demonstrates practical efficiency suitable for mobile deployment. Feature extraction processing time averages 4.3 seconds per image on standard smartphone hardware, while SVM classification requires less than 4 seconds once features are computed. The complete pipeline from image capture to pig identification takes approximately 8.3 seconds, which is feasible for real-time farm applications where immediate identification results are valuable for management decisions.




\subsection{Comparison with State-of-the-Art auricular vein based pig identification methods }
In this section, we compare our proposed approach with two state-of-the-art auricular vein-based pig identification methods in terms of both methodology and performance, as summarized in Table~\ref{tab:methodology_comparison} and Table~\ref{tab:literature_comparison}, respectively.



Dan et al. \cite{b16} employed HOG descriptors, which effectively capture local gradient orientations and edge information. Although this approach achieved a high accuracy of 98.50\%, it was evaluated on a limited data set of 5 Ghungroo pigs under controlled conditions. In addition, it relied on gradient-based features, which, despite their effectiveness in capturing edge information, can be sensitive to variations in lighting and noise. 

Similarly, Dan et al. \cite{9782062} focused on branching point templates with the help of Euclidean distance matching for vein branch point identification. This approach captured topological information about bifurcation locations but does not consider other significant biometric information such as endpoint distributions, inter-branch distances, angular relations, and global vein network geometry. Their 98.18\% accurate classification of 54 Yorkshire pigs, although exceptional by single-breed standards, highlights the limitations of topology-only feature extraction when addressing structurally diverse populations, where breed-specific variations in vein patterns compromise template-based matching effectiveness.


\begin{table}[t]
\centering
\caption{Methodological advantages of the proposed method over state-of-the-Art auricular vein based pig identification methods.}
\label{tab:methodology_comparison}
\resizebox{\textwidth}{!}{
\begin{tabular}{|p{4.1cm}|p{2cm}|C{3.48cm}|C{3.2cm}|}
\hline
\textbf{Aspect} & \textbf{AEIN\cite{9782062}} & \textbf{Black Pig SVM \cite{b16}} & \textbf{Proposed Method}\\
\hline
Breed diversity & Single & Single & Multiple\\
\hline
Feature comprehensiveness & Branching & HOG blocks & 68D vein analysis\\
\hline
Scalability test & Not reported & 5 pigs & 20 pigs\\
\hline
Robustness validation & Single breed & Single breed & Multiple breed\\
\hline
Feature dimension & Variable & HOG-dependent & 68D standardized\\
\hline
Generalization & Limited & Not tested & Cross-validated \\
\hline
\end{tabular}
}
\end{table}




\begin{table}[t]
\caption{Performance comparison with state-of-the-Art auricular vein based pig identification methods demonstrating competitive accuracy on mixed-breed datasets.}
\label{tab:literature_comparison}
\resizebox{\textwidth}{!}{
\begin{tabular}{|p{3cm}| C{2.3cm}| C{5cm}| C{2.5cm}|}
\hline
\textbf{Approach} & \textbf{Dataset size} & \textbf{Breed type} & \textbf{Accuracy (\%)}\\
\hline
Black Pig SVM \cite{b16}& 54 pigs & Single:Yorkshire & 98.18\\
\hline
AEIN\cite{9782062} & 5 pigs & Single: Ghungroo & 98.50\\
\hline
Proposed Method & 20 pigs & Mixed:  Landrace--Pietrain and Duroc--Pietrain & 98.12\\
\hline
\end{tabular}
}
\end{table}
Our 68-dimensional feature representation addresses the limitations of the above-mentioned methods by integrating multiple complementary aspects of the characteristics of the auricular vein. Specifically, \emph{it captures structural information via endpoint and bifurcation statistics, spatial positioning through coordinate analysis and density mapping, and geometric relationships using angular histograms.}  This multimodal representation enhances the discriminative power of our approach, making it robust to breed variability, as demonstrated by the accuracy achieved: \emph{98. 12\% in mixed-breed populations} ---a challenging scenario that was not considered in earlier studies. 

The reduction in accuracy (0.06–0.38\%) compared to single breed experiments reflects the added complexity of distinguishing between multiple breeds, where morphological differences introduce additional classification challenges. However, this modest performance drop suggests that our comprehensive feature extraction approach remains robust in managing breed variability, an area where more constrained methods, such as those relying solely on HOG descriptors or branching point templates, may face greater limitations.
 
\section{Conclusions and future work}
\label{sec:conclusions}
In this paper, we presented a cross-breed pig identification approach based on the unique patterns of the auricular veins. The proposed approach includes a comprehensive computer vision pipeline that extracted a 68-dimensional feature vector, effectively capturing the structural, spatial, and geometric characteristics of vein networks. This pipeline allows to isolate complex vascular structures within the ear, producing clear vein patterns suitable for detailed feature analysis.

Building on this, we introduced a lightweight identification system specifically designed for cross-breed pig populations. The system demonstrated high performance, achieving 98.12\% precision in mixed populations, addressing a critical gap in current state-of-the-art methods. Beyond its accuracy, the system was optimized for computational efficiency, and the entire identification pipeline was completed in just 8.3 seconds on standard smartphone hardware.

Beyond its technical performance, the system demonstrated significant potential for real-world impact. By relying only on low-cost smartphone cameras and simple lighting setups, barriers related to specialized equipment, ongoing costs, and technical expertise are removed. Although currently focused on identification, this capability lays the foundation for future applications in precision livestock management, offering resource-constrained farmers promising opportunities to improve health monitoring, breeding records, and overall productivity.


Although the system demonstrates strong performance, certain limitations remain.
The few misclassifications observed (1.88\% error rate) highlight opportunities for improvement. Future work will explore temporal vein pattern analysis to assess stability over time, potentially enabling change detection for health monitoring applications. Integration with complementary biometric modalities, such as facial features or body markings, through multimodal fusion could further enhance precision and provide redundancy for challenging identification scenarios. Furthermore, expanding the data set to include larger populations, additional breed combinations, and extended age ranges would strengthen generalization claims.


\section{Acknowledgment}
This work was partially supported by the Upanzi DPI Network at Carnegie Mellon University Africa.

\section{Data availability}
The dataset is available upon reasonable request from the corresponding author for academic research purposes, subject to appropriate ethical approval and data use agreements.
\bibliographystyle{elsarticle-num}

\bibliography{Reference}

\begin{thebibliography}{10}
\expandafter\ifx\csname url\endcsname\relax
  \def\url#1{\texttt{#1}}\fi
\expandafter\ifx\csname urlprefix\endcsname\relax\def\urlprefix{URL }\fi
\expandafter\ifx\csname href\endcsname\relax
  \def\href#1#2{#2} \def\path#1{#1}\fi

\bibitem{kapun2020uhf}
A.~Kapun, F.~Adrion, E.~Gallmann, Case study on recording pigs' daily activity patterns with a uhf-rfid system, Agriculture 10~(11) (2020) 542.
\newblock \href {https://doi.org/10.3390/agriculture10110542} {\path{doi:10.3390/agriculture10110542}}.

\bibitem{fuentes2022ai}
S.~Fuentes, C.~Gonzalez~Viejo, E.~Tongson, F.~R. Dunshea, The livestock farming digital transformation: Implementation of new and emerging technologies using artificial intelligence, Animal Health Research Reviews 23~(1) (2022) 59--71.
\newblock \href {https://doi.org/10.1017/S1466252321000177} {\path{doi:10.1017/S1466252321000177}}.

\bibitem{meng2025biometrics}
H.~Meng, L.~Zhang, F.~Yang, L.~Hai, Y.~Wei, L.~Zhu, et~al., Livestock biometrics identification using computer vision approaches: A review, Agriculture 15~(1) (2025) 102.
\newblock \href {https://doi.org/10.3390/agriculture15010102} {\path{doi:10.3390/agriculture15010102}}.

\bibitem{neethirajan2023digital2030}
S.~Neethirajan, \href{https://www.researchgate.net/publication/376798983_Digital_Livestock_Farming_2030_and_Beyond}{Digital livestock farming 2030 and beyond}, in: Proceedings of the International Conference on Systems and Technologies for Smart Agriculture (ICSTA 2023), 2023, conference paper, preprint available on ResearchGate.
\newline\urlprefix\url{https://www.researchgate.net/publication/376798983_Digital_Livestock_Farming_2030_and_Beyond}

\bibitem{himu2024digital}
H.~A. Himu, A.~Raihan, \href{https://www.researchgate.net/publication/384940499}{Digital transformation of livestock farming for sustainable development}, International Journal of Livestock Research 14~(9) (2024) 1--11, accessed July 2, 2025.
\newline\urlprefix\url{https://www.researchgate.net/publication/384940499}

\bibitem{dan2023ear}
S.~Dan, S.~Das, S.~Nath~Mandal, S.~Mustafi, S.~Banik, Ear image-based individual pig identification by using statistical parameters, Applied Biology Research 25~(1) (2023) 62--70.
\newblock \href {https://doi.org/10.5958/0974-4517.2023.00007.1} {\path{doi:10.5958/0974-4517.2023.00007.1}}.

\bibitem{mustafi2020biometric}
S.~Mustafi, P.~Ghosh, S.~Dan, K.~Mukherjee, K.~Roy, S.~Nath~Mandal, Biometrics-based pig identification: From invention to commercialisation, in: Lecture Notes in Electrical Engineering, Vol. 686, Springer, 2020, pp. 63--75.
\newblock \href {https://doi.org/10.1007/978-981-15-7031-5_7} {\path{doi:10.1007/978-981-15-7031-5_7}}.

\bibitem{ziyabari2014rfid}
S.~S.~K. Ziyabari, I.~Aris, \href{https://www.jnsciences.org/agri-biotech/20-volume-12/44}{A critical review of sustainable radio frequency identification (rfid)-based livestock monitoring and management systems: Towards quality products and practices}, Journal of New Sciences 12, accessed: 2025-08-27 (dec 2014).
\newline\urlprefix\url{https://www.jnsciences.org/agri-biotech/20-volume-12/44}

\bibitem{peng2025ear}
W.~Peng, Z.~Liu, J.~Cai, Y.~Zhao, Research and application progress of electronic ear tags as infrastructure for precision livestock industry: A review, Intelligent Robotics 5~(2) (2025) 433--449.
\newblock \href {https://doi.org/10.20517/ir.2025.22} {\path{doi:10.20517/ir.2025.22}}.

\bibitem{lomax2018ear}
S.~Lomax, E.~Hall, L.~Oehlers, P.~White, Topical vapocoolant spray reduces nociceptive response to ear notching in neonatal piglets, Veterinary Anaesthesia and Analgesia 45~(3) (2018) 366--373.
\newblock \href {https://doi.org/10.1016/j.vaa.2016.08.012} {\path{doi:10.1016/j.vaa.2016.08.012}}.

\bibitem{rob2024rfid}
Rob, The benefits of using rfid technology in pig feeders, available at: \url{https://www.barnworld.com/feeders/the-benefits-of-using-rfid-technology-in-pig-feeders/} (Accessed: 2025-07-17) (apr 2024).

\bibitem{cihan2023biometric}
P.~Cihan, A.~Saygili, N.~Eren~Ozmen, M.~Akyuzlu, Identification and recognition of animals from biometric markers using computer vision approaches: A review, Kafkas Univ Vet Fak Derg (2023).
\newblock \href {https://doi.org/10.9775/kvfd.2023.30265} {\path{doi:10.9775/kvfd.2023.30265}}.

\bibitem{10748293}
S.~Xu, G.~Liu, Y.~Liu, Y.~Wang, Research and application of pig iris recognition technology, in: 2024 3rd International Conference on Artificial Intelligence, Internet of Things and Cloud Computing Technology (AIoTC), 2024, pp. 1--7.
\newblock \href {https://doi.org/10.1109/AIoTC63215.2024.10748293} {\path{doi:10.1109/AIoTC63215.2024.10748293}}.

\bibitem{b15}
G.~N. Kimani, P.~Oluwadara, P.~Fashingabo, M.~Busogi, E.~Luhanga, K.~Sowon, et~al., Cattle identification using muzzle images and deep learning techniques, \url{https://arxiv.org/abs/2311.08148v1}, accessed: Jun. 11, 2024 (2024).

\bibitem{chakraborty2020pig}
S.~Chakraborty, et~al., Investigation on the muzzle of a pig as a biometric for breed identification, in: B.~B. Chaudhuri, M.~Nakagawa, P.~Khanna, S.~Kumar (Eds.), Proceedings of 3rd International Conference on Computer Vision and Image Processing, Springer, Singapore, 2020, pp. 71--83.
\newblock \href {https://doi.org/10.1007/978-981-32-9088-4_7} {\path{doi:10.1007/978-981-32-9088-4_7}}.

\bibitem{Rong2023facial}
W.~Rong, et~al., Pig face recognition based on metric learning by combining a residual network and attention mechanism, Agriculture 13~(1) (2023) 144.
\newblock \href {https://doi.org/10.3390/agriculture13010144} {\path{doi:10.3390/agriculture13010144}}.

\bibitem{b18}
R.~Ma, et~al., A lightweight pig face recognition method based on automatic detection and knowledge distillation, Applied Sciences 14~(1) (2024) 259.
\newblock \href {https://doi.org/10.3390/app14010259} {\path{doi:10.3390/app14010259}}.

\bibitem{b19}
J.~S. Yeon, R.~Ma, S.-C. Kim, Pig face recognition application using yolo algorithm and transformer model, in: Proceedings of the Future Technologies Conference (FTC) 2024, Vol. 795 of Lecture Notes in Networks and Systems, Springer, 2024, pp. 647--654.
\newblock \href {https://doi.org/10.1007/978-3-031-44851-5_52} {\path{doi:10.1007/978-3-031-44851-5_52}}.

\bibitem{9782062}
S.~Dan, et~al., Individual identification of black pig through ear images using support vector machine, in: 2021 4th International Conference on Recent Trends in Computer Science and Technology (ICRTCST), 2022, pp. 169--174.
\newblock \href {https://doi.org/10.1109/ICRTCST54752.2022.9782062} {\path{doi:10.1109/ICRTCST54752.2022.9782062}}.

\bibitem{dan2021ear}
S.~Dan, K.~Mukherjee, S.~Roy, S.~N. Mandal, D.~K. Hajra, S.~Banik, Individual pig recognition based on ear images, in: D.~Bhattacharjee, D.~K. Kole, N.~Dey, S.~Basu, D.~Plewczynski (Eds.), Proceedings of International Conference on Frontiers in Computing and Systems, Vol. 1255 of Advances in Intelligent Systems and Computing, Springer, Singapore, 2021.
\newblock \href {https://doi.org/10.1007/978-981-15-7834-2_55} {\path{doi:10.1007/978-981-15-7834-2_55}}.

\bibitem{b16}
S.~Dan, et~al., Aein-an intelligent computational technique for biometric based individual yorkshire pig identification using auricular vein, National Academy of Sciences Letters (2024) 1\href {https://doi.org/10.1007/s40009-024-01482-5} {\path{doi:10.1007/s40009-024-01482-5}}.

\bibitem{otsu_thresholding}
N.~Otsu, A threshold selection method from gray-level histograms, IEEE Transactions on Systems, Man, and Cybernetics 9~(1) (1979) 62--66.
\newblock \href {https://doi.org/10.1109/TSMC.1979.4310076} {\path{doi:10.1109/TSMC.1979.4310076}}.

\bibitem{MAKSYMENKO201795}
K.~Maksymenko, B.~Giusiano, N.~Roehri, C.-G. Bénar, J.-M. Badier, \href{https://www.sciencedirect.com/science/article/pii/S0165027017302583}{Strategies for statistical thresholding of source localization maps in magnetoencephalography and estimating source extent}, Journal of Neuroscience Methods 290 (2017) 95--104.
\newblock \href {https://doi.org/https://doi.org/10.1016/j.jneumeth.2017.07.015} {\path{doi:https://doi.org/10.1016/j.jneumeth.2017.07.015}}.
\newline\urlprefix\url{https://www.sciencedirect.com/science/article/pii/S0165027017302583}

\bibitem{gaussian_adaptive_thresholding}
N.~A. Rehman, F.~Haroon, \href{https://www.mdpi.com/2673-4591/32/1/23}{Adaptive gaussian and double thresholding for contour detection and character recognition of two-dimensional area using computer vision}, Engineering Proceedings 32~(1) (2023).
\newblock \href {https://doi.org/10.3390/engproc2023032023} {\path{doi:10.3390/engproc2023032023}}.
\newline\urlprefix\url{https://www.mdpi.com/2673-4591/32/1/23}

\bibitem{opencv_morph}
{OpenCV Team}, Morphological transformations --- opencv documentation, \url{https://docs.opencv.org/4.x/d9/d61/tutorial_py_morphological_ops.html}, accessed: 2025-08-25 (2023).

\bibitem{clahe}
S.~Pizer, R.~Johnston, J.~Ericksen, B.~Yankaskas, K.~Muller, Contrast-limited adaptive histogram equalization: speed and effectiveness, in: [1990] Proceedings of the First Conference on Visualization in Biomedical Computing, 1990, pp. 337--345.
\newblock \href {https://doi.org/10.1109/VBC.1990.109340} {\path{doi:10.1109/VBC.1990.109340}}.

\bibitem{canny}
H.~Agrawal, K.~Desai, Canny edge detection: A comprehensive review, International Journal of Technical Research \& Science 9 (2024) 27--35.
\newblock \href {https://doi.org/10.30780/specialissue-ISET-2024/023} {\path{doi:10.30780/specialissue-ISET-2024/023}}.

\bibitem{Davies2003VisualInspection}
E.~R. Davies, \href{https://www.sciencedirect.com/science/article/pii/B0122274105006633}{Visual inspection, automatic (robotics)}, in: R.~A. Meyers (Ed.), Encyclopedia of Physical Science and Technology, 3rd Edition, Academic Press, 2003, pp. 489--508.
\newblock \href {https://doi.org/10.1016/B0-12-227410-5/00663-3} {\path{doi:10.1016/B0-12-227410-5/00663-3}}.
\newline\urlprefix\url{https://www.sciencedirect.com/science/article/pii/B0122274105006633}

\bibitem{friedrich2023regularization}
S.~Friedrich, et~al., Regularization approaches in clinical biostatistics: A review of methods and their applications, Statistical Methods in Medical Research 32~(2) (2023) 425--440.
\newblock \href {https://doi.org/10.1177/09622802221133557} {\path{doi:10.1177/09622802221133557}}.

\end{thebibliography}
\end{document}